\documentclass{article}
\usepackage{neurips_2022}

\usepackage[utf8]{inputenc} % allow utf-8 input
\usepackage[T1]{fontenc}    % use 8-bit T1 fonts
\usepackage{hyperref}       % hyperlinks
\usepackage{url}            % simple URL typesetting
\usepackage{booktabs}       % professional-quality tables
\usepackage{amsfonts}       % blackboard math symbols
\usepackage{nicefrac}       % compact symbols for 1/2, etc.
\usepackage{microtype}      % microtypography
\usepackage{subfiles}

\usepackage{graphicx}
\usepackage{amsmath}
\usepackage{amssymb}
\usepackage{booktabs}
\usepackage{gensymb}
\usepackage[utf8]{inputenc} % allow utf-8 input
\usepackage[T1]{fontenc}    % use 8-bit T1 fonts
\usepackage{url}            % simple URL typesetting
\usepackage{booktabs}       % professional-quality tables
\usepackage{amsfonts}       % blackboard math symbols
\usepackage{nicefrac}       % compact symbols for 1/2, etc.
\usepackage{microtype}      % microtypography
\usepackage{lipsum}
\usepackage{fancyhdr}       % header
\usepackage{graphicx}       % graphics
\graphicspath{{media/}}     % organize your images and other figures under media/ folder
\usepackage{floatrow}
\usepackage{array}% http://ctan.org/pkg/array
\usepackage[HTML]{xcolor}
\usepackage{authblk}

\title{\hrule height 4pt
  \vskip 0.25in
  \vskip -\parskip\bf Unsupervised Discovery of Semantic Concepts in Satellite Imagery \\with Style-based Wavelet-driven Generative Models\vskip 0.29in
  \vskip -\parskip
  \hrule height 1pt
  \vskip 0.09in}

\author[1]{Nikos Kostagiolas}
\author[1]{Mihalis A. Nicolaou}
\author[2]{Yannis Panagakis}
\affil[1]{The Cyprus Institute}
\affil[2]{National and Kapodistrian University of Athens}

\date{}

\begin{document}

\maketitle

\begin{abstract}
  In recent years, considerable advancements have been made in the area of Generative Adversarial Networks (GANs), particularly with the advent of
   style-based architectures that address many key shortcomings - both in terms of modeling capabilities and network interpretability. Despite these improvements, the adoption of such approaches in the domain of satellite imagery is not straightforward.  Typical vision datasets used in generative tasks  are  well-aligned and annotated, and exhibit limited variability.  In contrast, satellite imagery exhibits great spatial and spectral variability, wide presence of fine, high-frequency details, while the tedious nature of annotating satellite imagery leads to annotation scarcity - further motivating developments in unsupervised learning.  In this light, we present the first pre-trained style- and wavelet-based GAN model\footnote{both the pre-trained model and our code will be made publicly available upon paper acceptance} that can readily synthesize a wide gamut of realistic satellite images in a variety of settings and conditions  - while also preserving high-frequency information.   Furthermore, we show that by analyzing the intermediate activations of our network, one can discover a multitude of interpretable { semantic directions} that facilitate the guided synthesis of satellite images in terms of high-level concepts (e.g., urbanization) without using any form of supervision.  Via a set of qualitative and quantitative experiments we demonstrate the efficacy of our framework, in terms of suitability for downstream tasks (e.g., data augmentation),  quality of synthetic imagery, as well as generalization capabilities to unseen datasets.
\end{abstract}

\begin{figure}[h]
\centering
\includegraphics[angle=90,width=1\linewidth]{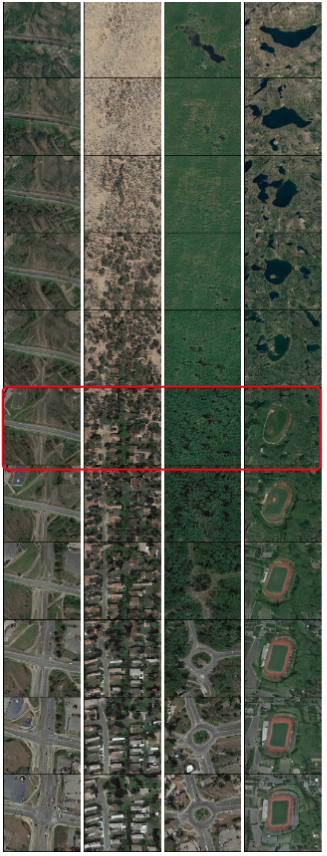}
\caption{Example of edits moving along the first semantic direction. Exploration begins from the examples in the sixth column (highlighted with red). Moving along this direction provides a significant change in urbanization levels as more and more human structures are getting introduced depending on the edit magnitude (columns 7 to 11). It is worth noting that our method is independent of background texture, as can seen by the various types of vegetation in the initial images (freeway with limited surrounding flora, chaparral, meadow/forest and wetland with a baseball court in rows 1, 2, 3 and 4 respectively, counting from bottom).}
\label{fig:introduction}
\end{figure}

\section{Introduction}

Synthetic image generation lies at the forefront of current deep learning research, owing largely to the fact that it forms a powerful and versatile medium for solving a multitude of image related tasks \cite{karras2019stylebased,ledig2017photorealistic,radford2016unsupervised,thies2019deferred}. Thanks to its outstanding breakthroughs, tasks like ill-posed inverse problems \cite{ongie2020deep}, image in-painting \cite{elharrouss2020image}, novel view synthesis \cite{unknown} and manipulation of specific image properties \cite{lee2020maskgan} have recently experienced a rapid surge in their performance, as more and more advancements both in visual quality and adaptability across different data modalities are being made by the day. This tremendous success has recently led to findings that indicate that GANs are able to capture multiple interpretable attributes in their latent space \cite{goetschalckx2019ganalyze,jahanian2020steerability,shen2020interpreting,plumerault2020controlling,yang2020semantic}, such as  facial attributes \cite{shen2020interpreting} and scene lighting conditions \cite{yang2020semantic}. The identification of such meaningful semantic directions paves the way towards establishing robust methods for guided control of such attributes, thus allowing for a vast selection of different applications ranging from facial attribute manipulation \cite{shen2020interpreting,gu2020image} to scene editing \cite{yang2020semantic,zhu2020unpaired}. 

Such developments are, however, not straightforwardly transferable to the domain of satellite imagery \cite{9351574}, and while there has been progress, it is not on par with results on other domains.  This can be largely attributed to the particularities of satellite image datasets, that are indeed much more complex - both in terms of structure as well as variability of objects and their attributes.  In order to capture the highly hierarchical, high-frequency content distribution \cite{cheng2020remote} of satellite images, data-intensive, deep learning models have in the past resorted to multi-modal approaches that involved adding varying levels of supervision to the pipeline \cite{deng2018like,10.1145/3209811.3212707,data_augment}, the performance of which relies heavily on the existence of reliable data collection and annotation processes. However, considering the extreme variability of satellite imagery data such processes tend to be rather strenuous, especially when data-intensive, deep learning pipelines are concerned\cite{9351574}. 

Progress towards strengthening the performance of generative networks, both in terms of their data and computational requirements as well as the quality and fidelity of their generations, has come in the form of progressive, style-based \cite{karras2018progressive,karras2019stylebased,karras2020analyzing} and, quite recently, frequency-based \cite{gal2021swagan} architectures. Especially the latter have been shown to tackle the shortcomings of the former in terms of fine, high-frequency detail modeling in a  robust manner.

In this work, we leverage the recent advancements in frequency-aware generative models in tandem with recent works on GAN semantic discovery \cite{shen2021closedform,peebles2020hessian,wu2020stylespace}  aiming to narrow the gap between state-of-the-art approaches in image generation and semantic exploration, and satellite imagery and manipulation. Concretely, 
%In this work, 
we present, to the best of our knowledge, the first Style-based GAN that can readily generate a multitude of realistic satellite imagery, leveraging recent advancements in frequency-aware generative models.  Furthermore, we show that the latent space of our model is enriched with semantic concepts that facilitate the guided synthesis of realistic satellite images in terms of semantic concepts, such as {\it urbanization} (Fig. \ref{fig:introduction}).  Finally, we show that synthesized data can be readily used in downstream tasks, such as data augmentation, improving results even in unseen datasets. In summary, our key contributions are as follows:
\begin{itemize}
    \item We present the first pre-trained Style-based  GAN for satellite image generation, that can well-capture high-frequency details inherent to satellite images, and can produce a wide gamut of imagery that includes different morphology, objects, and land-use types.   
    \item We show that unsupervised analysis of the latent space of our GAN results in distinct semantic directions that correspond to high-level concepts (such as {\it urbanization, vegetation growth, wetland density}), and facilitates guided editing for controllable synthesis.  This is to the best of our knowledge, the first attempt to discover such concepts in the domain of satellite imagery.
    \item By a set of qualitative and quantitative results on several datasets (NWPU-RESISC45, AID and UC-Merced), we demonstrate both the quality of synthesized results, as well as the suitability for use in downstream tasks. Concretely, we show that data augmentation using our pre-trained model can result in increased accuracy on tasks such as remote scene classification.  Interestingly, we show that this approach generalizes to unseen data; that is, we can augment datasets on which the generative model has not been trained on (e.g., UC-Merced and AID) - and still improve performance.
\end{itemize}
\label{sec:intro}

\section{Related Work}
\label{sec:related}

\subsection{Generative Adversarial Networks}
Generative Adversarial Networks (GANs), introduced by Goodfellow et al. \cite{goodfellow2014generative}, continue to provide impressive results over the recent years, with a multitude of applications in synthesis-related tasks (e.g.,  image editing \cite{richardson2021encoding,suzuki2019spatially,park2020swapping,gu2020image,menon2020pulse,alaluf2021matter}, domain translation \cite{zhu2020unpaired,liu2018unsupervised,wang2018highresolution,park2019semantic} and video generation \cite{tulyakov2017mocogan,2020,hao2021gancraft}).  Recently, further improvements in the fidelity of GAN-based generation have been brought on via novel architecture variants for GANs, such as progressive-growing \cite{karras2018progressive} and style-based architectures \cite{karras2019stylebased,karras2020analyzing}.  By architectural changes such as the introduction of Adaptive Instance Normalization and a mapping network that maps a random sample to an intermediate latent space, style-based architectures have become the go-to models for high-fidelity image synthesis. 

Another element that has been recently introduced into the GAN community is to augment the model with spectral information, that can lead to better modeling of high-frequency content.
Indeed, a significant number of studies in the convolutional neural network  \cite{7532596,liu2019multilevel,waveletpooling,mildenhall2020nerf,tancik2020fourier} and the generative domain alike \cite{article,wang2020multilevel,10.1145/3357254.3358600,liu2019attributeaware} have stressed the benefits of modeling frequency-based information towards alleviating issues that arise due to spectral bias.  An end-to-end frequency-based image generation model has recently been presented in Gal et al. \cite{gal2021swagan}, where content is directly generated in wavelet-space, utilizing the inverse wavelet transform for mapping back into image space.  Although evaluation was conducted on relatively homogeneous datasets (namely the FFHQ \cite{karras2019stylebased}, LSUN \cite{yu2016lsun} and CelebA \cite{liu2015faceattributes}), the wavelet-driven approach adopted in \cite{gal2021swagan} can also be suitable for the task of modelling the distribution of real satellite imagery.

\subsubsection{Satellite Imagery Generation}
The majority of machine learning applications in the domain of satellite images often relies on using  costly annotations, that are often scarce due to their strenuous collection process, or even plagued by lack of sufficient quality.  At the same time, fine-tuning pre-trained deep architectures trained on common vision datasets is not sufficient for capturing the large variability of satellite images \cite{Survey}.

Although such constraints formed the main target of past studies that employed transfer learning \cite{10.1145/3209811.3212707} or data augmentation techniques \cite{data_augment} to tackle them, the resulting models lacked generalization capabilities \cite{article2,inproceedings}. An alternative to those under-performing approaches quickly surfaced in the form of unsupervised generation \cite{2017,2019,edge_enhanced,duan2018gan,bashmal2018siamese,xu2018remote,teng2019classifier,han2018semi,yu2019attention} mainly tailored for super-resolution, data augmentation, and adversarial feature learning strategies for scene classification applications. Nevertheless, despite the successful  generation of satellite-like imagery, the overall results were either i) plagued by the presence of heavy noise and lack of semantic disentanglement which rendered the structures present in the resulting generations highly indiscernible and of limited realism \cite{Marta} or ii) even in cases in which it was suggested that authentic images were able to be generated, the resulting performance was generally inferior to mainstream CNN-based approaches \cite{Survey}. This indicates that, although CNN-based approaches have been generally constrained to traditional geometric-based data augmentation methods for enhancing their classification performance, any trials at incorporating context-based variability to the overall pipeline by the adoption of auxiliary generative models have yet to be explored.

\subsection{Latent Semantic Exploration}
Recent results have showcased the ability of GAN-like architectures and variants to encode a variety of semantics  in the intermediate feature  \cite{bau2018gan} and  latent space \cite{goetschalckx2019ganalyze,jahanian2020steerability,shen2020interpreting,yang2020semantic}. However, distinguishing such semantics in an implicit manner has had, for the most part, relied on supervised learning techniques that require training a classifier given a collection of human-labeled generation samples, a process which introduces biases to the whole pipeline. Several unsupervised approaches opted to bypass that limitation \cite{harkonen2020ganspace,voynov2020unsupervised} albeit still partially entailing model training and data sampling; nevertheless, later, purely unsupervised approaches were proposed \cite{shen2021closedform,peebles2020hessian,wu2020stylespace} that managed to successfully detect interpretable directions while being proof of any training or sampling. This was achieved by either employing eigenvector decomposition and treating those eigenvectors associated with the highest eigenvalues as manipulation directions \cite{shen2021closedform}, using Hessian loss minimization \cite{peebles2020hessian} or identifying localized controls by semantic mapping \cite{wu2020stylespace}. Even though those models still require substantial human effort, the interest in closed-form latent semantic discovery has, as of late, surged owing to its potential applicability in a wide range of applications.

\section{Approach}
\label{sec:approach}

Our goal for this present study is twofold: first, to train an image synthesis model in a fully unsupervised way able to reliably generate accurate, real-like RGB satellite images while also covering a wide variety of plausible context/background combinations and, secondly but no less importantly, to identify meaningful semantic directions within the trained model's latent space.

\subsection{Generative Model}
Due to the  hierarchical nature present in satellite imagery data across scales, we follow the line of work that adopts style-based GAN architectures, which injects style information across layers via the AdaIN operator.  Thus style information modulates the image generation process across different scales in a hierarchical manner.  Furthermore, in order to leverage the rich spectral content that accompanies the diverse nature of satellite imagery, we adopt the SWAGAN architecture \cite{gal2021swagan} that introduces wavelet decompositions in the layers of StyleGAN. 

The SWAGAN model first decomposes a given image by mapping it into high-frequency space and then proceeds to progressively downscale it until it has acquired a low-frequency representation. Then, via a series of upscaling transformations, high-frequency content can be generated from the  low-frequency latent space. Every upscaling and downscaling process relies on wavelet and inverse wavelet transforms, respectively. This architecture has been shown to alleviate issues arising due to spectral bias of GANs\cite{gal2021swagan}, and can hence  handle high-frequency content (i.e., image details) in a more reliable manner.  

Our SWAGAN model was trained and evaluated exclusively on the RESICS45 dataset with a batch size of $64$ on four NVIDIA Tesla V100-SXM2 32GB. For this purpose, our SWAGAN model was trained on the RESICS45 dataset for $200,000$ iterations.
Our codebase is based on the PyTorch StyleGAN2 implementation\footnote{https://github.com/rosinality/stylegan2-pytorch} and the official PyTorch SeFa implementation\footnote{https://github.com/genforce/sefa} with minor porting adjustments.

\subsection{Unsupervised Discovery of Semantic Concepts}
We aim to discover high-level semantic concepts that arise in satellite imagery, such as urbanization, vegetation growth, and wetland density.  Discovering such concepts  facilitates the controllable generation of satellite images, and can therefore be utilized for downstream tasks such as data augmentation.

Recent work has shown that semantic concepts are encapsulated in the latent space of GANs \cite{shen2021closedform,voynov2020unsupervised}.  In order to discover such concepts in an unsupervised manner, one can apply Principal Component Analysis (PCA) on the weight matrix $\mathbf{A}$ of a projection layer  of the generative model, thus finding the directions that lead to large variation in the output images.    As shown in \cite{shen2021closedform}, these directions correspond to the top-$k$ eigenvectors $\mathbf{u}_i$ (corresponding in turn to the $k$ eigenvalues $n_i$)  of the covariance matrix of weights, $\mathbf{A}^T\mathbf{A}$.  

Having discovered a set of interpretable directions, one can perform an edit by manipulating the latent code $\mathbf{z}$ in the direction $\mathbf{u}_i$ corresponding to the desired concept by $\mathbf{z}' = \mathbf{z}+\alpha\mathbf{u}_i$, where $\alpha$ is a scalar that corresponds to the edit magnitude. The edited image $\mathbf{y}'$ can  be obtained by 
\begin{equation}
\mathbf{y}' \triangleq G(\mathbf{z}')=G( \mathbf{z}+\alpha\mathbf{u}_i) ,  
\end{equation}
where $G$ denotes the application of the SWAGAN generator.

\begin{figure*}
  \centering
  \begin{tabular}{ m{2cm} m{2cm} m{2cm} m{2cm} m{2cm} m{2cm} m{2cm}}
   \centering
   \includegraphics[width=\linewidth]{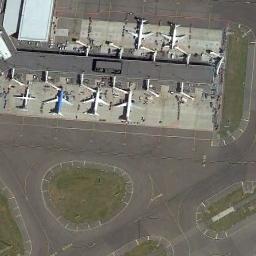}
   \small (a) {Airplane}
   &
   \includegraphics[width=\linewidth]{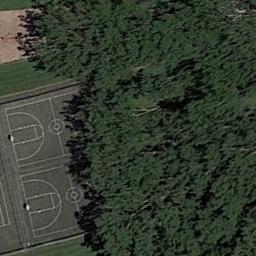}
   \small (b) {Bask. Court}
   &
   \includegraphics[width=\linewidth]{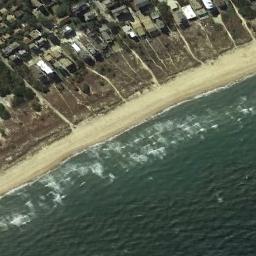}
   \small (c) {Beach}
   &
   \includegraphics[width=\linewidth]{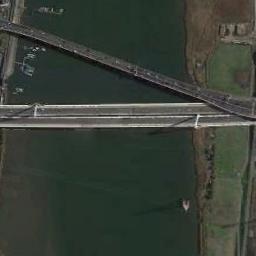}
   \small (d) {Bridge}
   &
   \includegraphics[width=\linewidth]{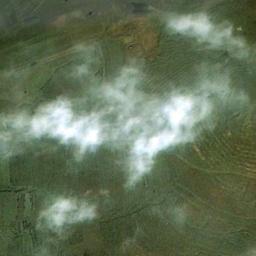}
   \small (e) {Cloud}
   &
   \includegraphics[width=\linewidth]{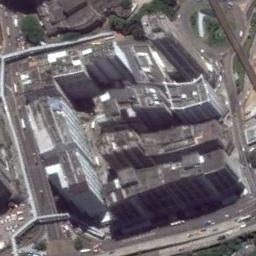}
   \small (f) {Comm. Area}
   &
   \includegraphics[width=\linewidth]{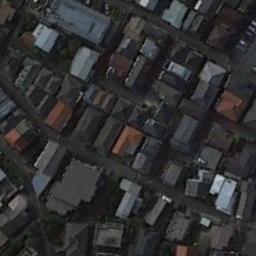}
   \small (g) {Dense Res.}\\
   \includegraphics[width=\linewidth]{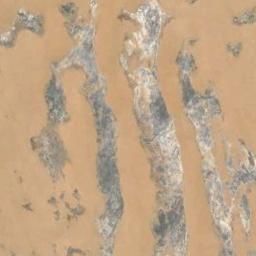}
   \small (h) {Desert}
   &
   \includegraphics[width=\linewidth]{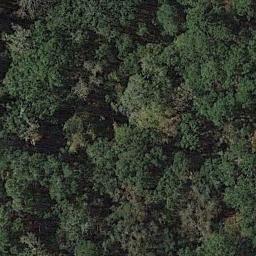}
   \small (i) {Forest}
   &
   \includegraphics[width=\linewidth]{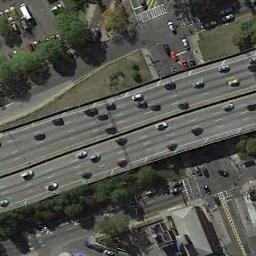}
   \small (j) {Freeway}
   &
   \includegraphics[width=\linewidth]{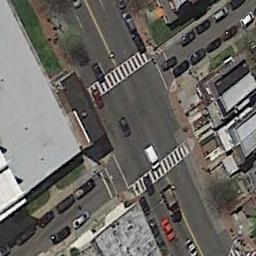}
   \small (k) {Intersection}
   &
   \includegraphics[width=\linewidth]{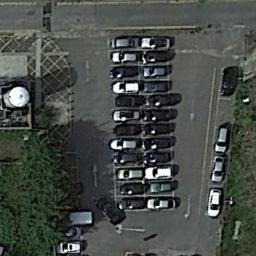}
   \small (l) {Parking Lot}
   &
   \includegraphics[width=\linewidth]{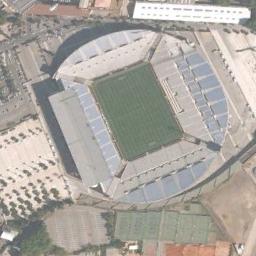}
   \small (m) {Stadium}
   &
   \includegraphics[width=\linewidth]{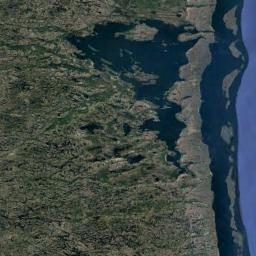}
   \small (n) {Wetland}
  \end{tabular}
  \caption{Example images from the RESICS45 dataset, class labels included.}
  \label{fig:resics}
\end{figure*}

\section{Experiments}
\label{sec:experiments}

In this section, we present a set of qualitative and quantitative experiments, that (i) demonstrate the quality and diversity of synthesized satellite images, (ii) showcase the high-level semantic concepts that are discovered, and the unsupervised editing that leads to controllable generation, and (iii) highlight the suitability of the proposed framework for tasks such as data augmentation for remote scene classification.  Interestingly, we find that via the proposed framework one can improve the accuracy on unseen data, by augmenting datasets the generative model has not been trained on. 

\subsection{Datasets}
\label{Datasets}
RESISC45 is a publicly available\footnote{https://www.tensorflow.org/datasets/catalog/resisc45} benchmark dataset for Remote Sensing Image Scene Classification (RESISC), created by the Northwestern Polytechnical University (NWPU). This dataset contains 31,500 optical satellite images of 256 x 256 resolution, covering 45 scene classes with 700 images in each class. A set of example images across every class can be browsed through in Fig. \ref{fig:resics}.

The AID dataset\cite{AID}, is collected from Google Earth, and comprises 10,000 600 × 600 images distributed across 30 classes. The class size varies from 220 to 420. 

The UC-Merced dataset\cite{Nilsback08}, consists of 2,100 256 × 256 overhead scene RGB images, evenly distributed across 21 classes. These images are selected from the United States Geological Survey (USGS) National Map.
\subsection{Realistic Optical Satellite Imagery Generation}
\label{Realistic}
\begin{figure*}
  \centering
  \includegraphics[width=0.9\textwidth]{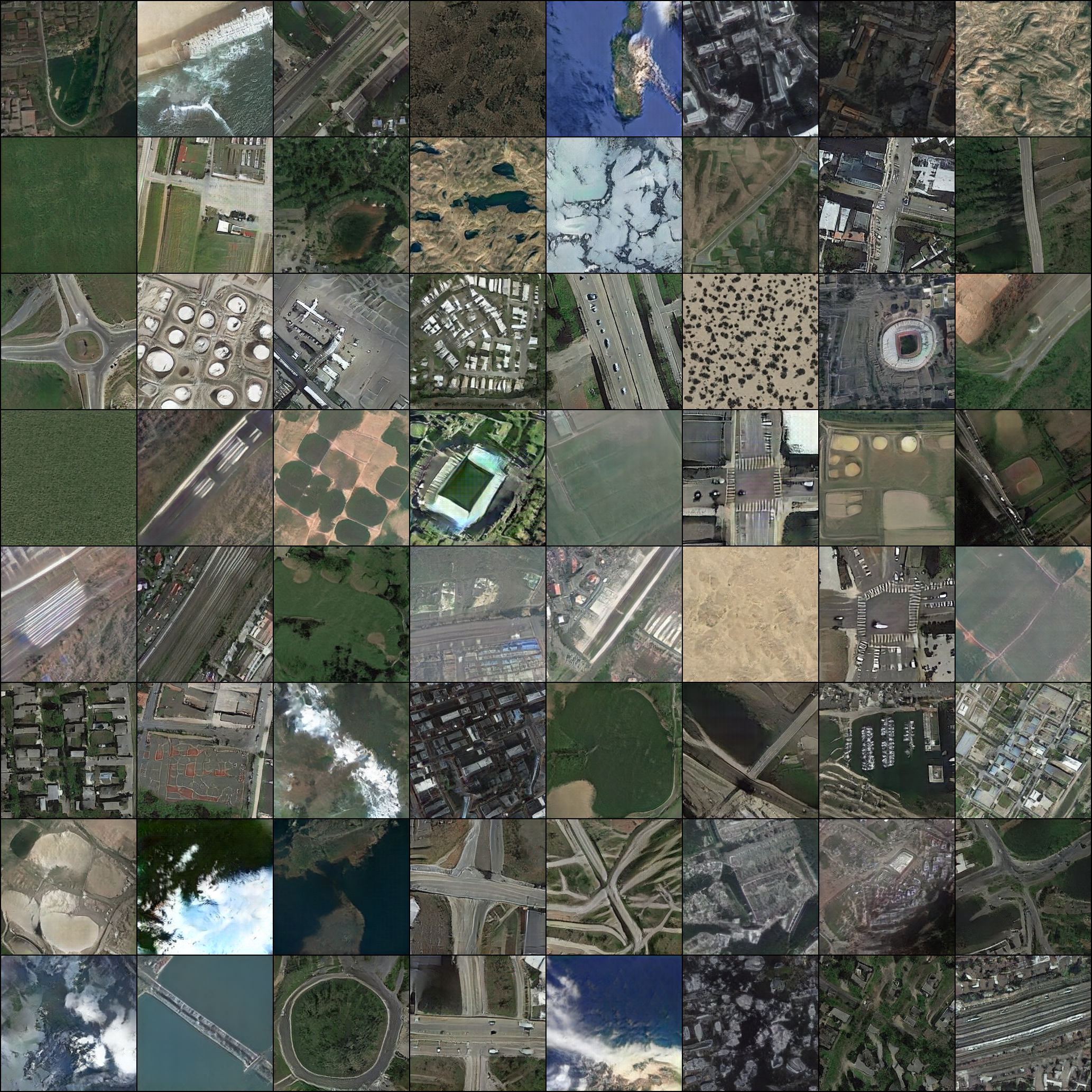}
  \caption{Uncurated satellite imagery samples generated by our SWAGAN model, using truncation with $\psi$ = 0.5. Zooming in is encouraged.}
  \label{fig:generations}
\end{figure*}
We first demonstrate the ability of our learned representation to efficiently capture high-frequency satellite imagery data with no supervision.  Figure \ref{fig:generations} displays uncurated output generations of the trained model. Even when dealing with the greatly varying patterns both structure- and texture-wise present in the data, our model is able to clearly capture them with great accuracy and plausibility.

Indeed, by visually inspecting the results, we can conclude that the generated images are realistic, and in most, if not all cases, indistinguishable from real ones.  Our pre-trained model is able to successfully capture and recreate even the finest details such as housing, vehicles, trees, rural roads and pathways, airplanes, and other objects, while also being versatile enough to capture a vast array of different spectral signatures, ranging from deserts to stadiums and car parking lots to circular farmlands and sea ice.  Furthermore, this is done in an unsupervised manner - in contrast to the majority of related work, where the synthesis of satellite imagery relies heavily on supervision and annotation availability \cite{Survey}.

\subsection{Controllable Synthesis of Satellite Imagery}
\label{Controllable}
\begin{figure*}
  \begin{tabular}{ m{3.7cm} m{3.7cm} m{3.7cm} m{3.7cm}}
   \includegraphics[width=1.1\linewidth]{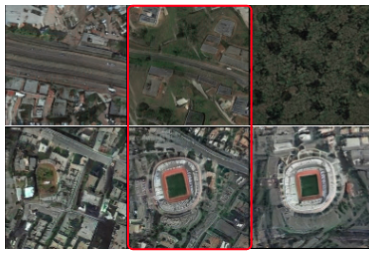}
   \centering\small (a)
   &
   \includegraphics[width=1.1\linewidth]{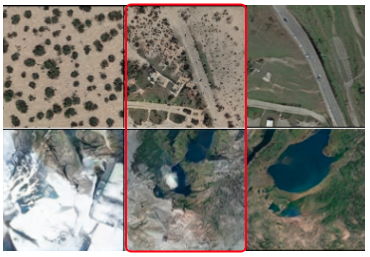}
   \centering\small (b)
   &
   \includegraphics[width=1.1\linewidth]{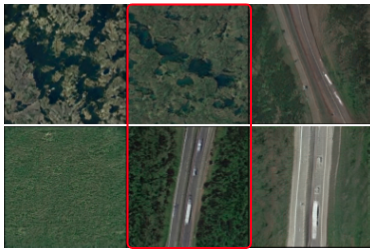}
   \centering\small (c)
   &
   \includegraphics[width=1.1\linewidth]{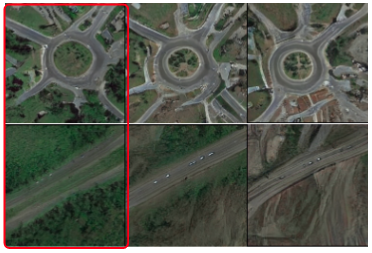}
   \centering\small (d)
   \end{tabular}
   \caption{Example of edits moving along the third (a), fourth (b), fifth (c), and seventh (d) semantic direction. Exploration begins from the examples in the column highlighted with red. (a) Moving along this direction reinforces structures along the vertical axis (stadium in bottom row), while moving away from it reinforces structure in the horizontal one (railway in top row). (b) Moving along this direction ups the presence of secondary structures within the image (road, lake, from top to bottom) while moving away from it reinforces the main texture (chaparral, mountain/iceberg, from top to bottom). (c) Moving along this direction introduces road structures within an image, while moving away from it completely removes them from the image. (d) Moving along this direction decreases flora presence while preserving the overall semantics almost in their entirety.}
   \label{fig:edits}
\end{figure*}

Our second series of experiments involves the discovery of interpretable directions in the latent space of our trained SWAGAN model.  In particular, we find that moving along the first principal component tends to affect  the overall semantic of an image, capturing the {\it urbanization} concept,  while edits along the less important directions seem to produce progressively lighter semantic alterations to the image as the eigenvalue of their corresponding eigenvectors decreases in magnitude (Figures \ref{fig:edits}a-\ref{fig:edits}d).  
Several directions that correspond to high-level concepts have been discovered by following this approach, with edits in the same direction having an identifiable, common effect on a large and diverse selection of images from different classes. 

\begin{figure*}
  \centering
  \begin{tabular}{ m{2cm} m{2cm} m{2cm} m{2cm} m{2cm} m{2cm} m{2cm}}
   \includegraphics[width=1.05\linewidth]{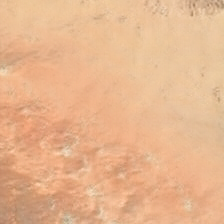}
   &
   \includegraphics[width=1.05\linewidth]{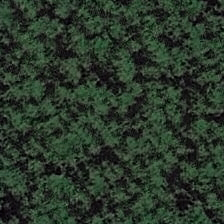}
   &
   \includegraphics[width=1.05\linewidth]{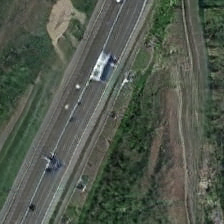}
   &
   \includegraphics[width=1.05\linewidth]{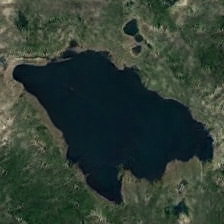}
   &
   \includegraphics[width=1.05\linewidth]{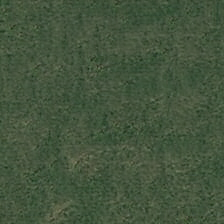}
   &
   \includegraphics[width=1.05\linewidth]{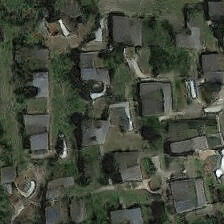}
   &
   \includegraphics[width=1.05\linewidth]{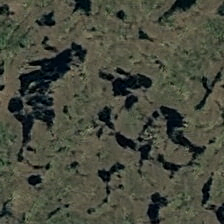}
  \end{tabular}
  \caption{Example generated images used to augment the NWPU-RESISC45 classes, starting from the left-most image: \textit{desert}, \textit{forest}, \textit{freeway}, \textit{lake}, \textit{meadow}, \textit{medium\_residential}, \textit{wetland}.}
  \label{fig:data_augmentations}
\end{figure*}

\begin{table*}[]
    \centering
    \begin{tabular}{*5l}    \toprule
    \emph{Semantic Direction \#} & \emph{Description (+)}  & \emph{Description (-)}\\\midrule
    1  & Urbanization Growth & Urbanization Diminishment\\ 
    3  & Vertical axis structure reinforcement & Horizontal axis structure reinforcement\\ 
    4  & Secondary structure growth & Main structure growth\\
    5  & Road structure introduction/reinforcement & Road structure removal\\
    7  & Flora diminishment upon existing structure & Flora growth upon existing structure\\\bottomrule
     \hline
    \end{tabular}
   \caption{Summary of correspondence between semantic direction number and edit description.}
    \label{tab:direction_summary}
\end{table*}

More specifically, manipulating the images with respect to the first direction allows us to control their urbanization levels, ranging from spectral signatures that can be assigned to the "meadow", "desert", and "wetland" classes (categories that entail little to no urbanization) all the way to classes such as "roundabout", "overpass" and "stadium" (where urbanization levels can be perceived as being, generally, at their peak). Moving along that direction can, depending on the magnitude of the edit action and the initial landscape composition, incorporate water level diminishment along with flora growth, a combination of flora diminishment with an increase in urbanization, or result in greater definition of already present urban structures. While not being as intervening as the first semantic edit direction, the following four identified types of semantic manipulation can be successfully matched with controls that seem to preserve the overall image textures with major or minor alterations in specific areas of focus. Namely, traversing the third and fourth semantic direction (Figures \ref{fig:edits}a and \ref{fig:edits}b, respectively) we can easily relate with definition-based alterations on structures that lie either on the vertical or horizontal axes of symmetry (Fig. \ref{fig:edits}a) and secondary structure reinforcement (Fig. \ref{fig:edits}b) while the fifth and seventh directions seem to be more-or-less involved with features of lesser scale such as introducing roads (Fig. \ref{fig:edits}c) and vegetation pruning (Fig. \ref{fig:edits}d).
Our findings with relation to semantic direction/description correspondence are summarized in Fig.  \ref{tab:direction_summary}.

\subsection{Data Augmentation for Remote Scene Classification}
\label{RemoteSceneClassification}
In order to further assess the suitability of the synthesized imagery for downstream tasks, we present a series of remote scene classification experiments applying data augmentation via our framework.  In particular, we utilize a set of well-known benchmark satellite imagery datasets,
which we artificially imbalance by reducing the available training samples within a variety of their classes, to make the setting more challenging. Interestingly, we find that data augmentation using our framework results in increased accuracy on tasks such as remote scene classification - even when applied on datasets that our generative model has not been trained on.  

For each of the datasets and settings considered, we compare the performance of a ResNet-50 neural network classifier separately trained on: (a) the imbalanced data (\textit{Imbalanced} in our experiments), (b) augmentation with baseline data  techniques (\textit{Baseline Data Aug.}), and (c) SeFA-based augmentation using our pre-trained SWAGAN, and (d) both baseline and SeFa (\textit{mixed}).

In particular, baseline techniques (b) constitute classical data augmentation approaches, relying on a set of affine transformations such as rotations and translations.  For (c), we augment the dataset using randomly generated samples stemming from our pre-trained SWAGAN, moving along the first semantic component, according to Sec. \ref{sec:approach}.

In essence the baseline data-augmentation approach introduces variability of a very different nature, in contrast to the edited images via our framework.  In particular, the baseline augmentation provides further geometric variability, while in our case concept-dependent variation is introduced - as can be inspected by editing results.  Hence, we expect that the contribution towards accuracy improvement from these two approaches can function in a {\it synergistic} manner.  This is verified by the experiments we present in subsection \ref{res_and_dis}, where adopting both augmentation methods further increases accuracy.

\subsubsection{ Dataset Variants Description}
We briefly summarize the dataset splits that have been used in our experiment.
\label{imbalanced_datasets}
\begin{itemize}
    \item \textbf{Resisc70}: In this imbalanced version of the NWPU Resisc-45 dataset we randomly sample 7 classes (from the overall 45) that are to be artificially imbalanced. While we maintain a 450/150/100 train/validation/test ratio for the balanced ones, the corresponding ratio for the imbalanced cases is set at 70/150/100.
    \item \textbf{Resisc35}: This version of the imbalanced NWPU Resisc-45 dataset is the same as the above, except for the fact that the available training examples for the imbalanced classes are reduced to 35.
    \item \textbf{Resisc10}: This version of the imbalanced NWPU Resisc-45 is the same as that of Resisc70, except for the fact that the available training examples for the imbalanced classes are limited to just 10.
    \item \textbf{Imbalanced UC Merced}: In this imbalanced version of the UC Merced dataset we randomly sample 5 classes (from the overall 21) that are to be artificially imbalanced. While the balanced ones have a 75/15/10 train/validation/test ratio, the corresponding ratio of the imbalanced cases lies at 10/15/10.
    \item \textbf{Imbalanced AID}: In this imbalanced version of the AID dataset we randomly sample 7 classes (from the overall 30) that are to be artificially imbalanced. While the balanced ones have a 120/40/40 train/validation/test ratio, the corresponding ratio of the imbalanced cases lies at 40/40/40.
\end{itemize}

\begin{table*}[h]
\small
\begin{tabular}{l|llllllll|}
\cline{2-9}
                                                    & \multicolumn{8}{c|}{\textbf{Balance Type}}                                                                                                                                                                                                                                                       \\ \cline{2-9} 
\textbf{}                                           & \multicolumn{2}{c|}{\textit{Imbalanced}}                                             & \multicolumn{2}{c|}{\textit{Baseline Data Aug.}}                                     & \multicolumn{2}{c|}{\textit{SeFa Data Aug.}}                                & \multicolumn{2}{c|}{\textit{Mixed}}                    \\ \hline
\multicolumn{1}{|l|}{\textbf{Dataset Variant}}                                          & \multicolumn{1}{l|}{\textbf{Acc.}} & \multicolumn{1}{l|}{\textbf{Imb.Acc.}} & \multicolumn{1}{l|}{\textbf{Acc.}} & \multicolumn{1}{l|}{\textbf{Imb.Acc.}} & \multicolumn{1}{l|}{\textbf{Acc.}} & \multicolumn{1}{l|}{\textbf{Imb.Acc.}} & \multicolumn{1}{l|}{\textbf{Acc.}} & \textbf{Imb.Acc.} \\ \hline
\multicolumn{1}{|l|}{\textit{Resisc-70}}            & \multicolumn{1}{l|}{0.827}              & \multicolumn{1}{l|}{0.638$\pm$0.191}                  & \multicolumn{1}{l|}{0.835}              & \multicolumn{1}{l|}{0.674$\pm$0.218}                  & \multicolumn{1}{l|}{\textbf{0.836}}              & \multicolumn{1}{l|}{\textbf{0.694$\pm$0.166}}                  & \multicolumn{1}{l|}{0.842*}              &       \multicolumn{1}{l|}{0.760$\pm$0.128*}            \\ \hline
\multicolumn{1}{|l|}{\textit{Resisc-35}}            & \multicolumn{1}{l|}{0.804}              & \multicolumn{1}{l|}{0.463$\pm$0.242}                  & \multicolumn{1}{l|}{\textbf{0.829}}              & \multicolumn{1}{l|}{\textbf{0.629$\pm$0.189}}                  & \multicolumn{1}{l|}{0.821}              & \multicolumn{1}{l|}{0.561$\pm$0.214}                  & \multicolumn{1}{l|}{0.833*}              &  \multicolumn{1}{l|}{0.673$\pm$0.137*}                 \\ \hline
\multicolumn{1}{|l|}{\textit{Resisc-10}}            & \multicolumn{1}{l|}{0.776}              & \multicolumn{1}{l|}{0.235$\pm$0.230}                  & \multicolumn{1}{l|}{0.764}              & \multicolumn{1}{l|}{0.270$\pm$0.256}                  & \multicolumn{1}{l|}{\textbf{0.786}}              & \multicolumn{1}{l|}{\textbf{0.370$\pm$0.248}}                  & \multicolumn{1}{l|}{0.808*}              &   \multicolumn{1}{l|}{0.494$\pm$0.222*}                \\ \hline
\multicolumn{1}{|l|}{\textit{Imb. UC-Merced}} & \multicolumn{1}{l|}{0.776}              & \multicolumn{1}{l|}{0.420$\pm$0.286}                  & \multicolumn{1}{l|}{0.810}              & \multicolumn{1}{l|}{0.580$\pm$0.319}                  & \multicolumn{1}{l|}{\textbf{0.814}}              & \multicolumn{1}{l|}{\textbf{0.600$\pm$0.303}}                  & \multicolumn{1}{l|}{0.819*}              &  \multicolumn{1}{l|}{0.660$\pm$0.356*}                 \\ \hline
\multicolumn{1}{|l|}{\textit{Imb. AID}}       & \multicolumn{1}{l|}{0.716}              & \multicolumn{1}{l|}{0.425$\pm$0.269}                  & \multicolumn{1}{l|}{0.766}              & \multicolumn{1}{l|}{\textbf{0.668$\pm$0.275}}                  & \multicolumn{1}{l|}{\textbf{0.767}}              & \multicolumn{1}{l|}{0.629$\pm$0.258}                  & \multicolumn{1}{l|}{0.783*}              &  \multicolumn{1}{l|}{0.732$\pm$0.176*}                 \\ \hline
\end{tabular}
\caption{Comparison between different data augmentation settings for 5 different dataset variants. We denote with \textit{Acc} the top-1 accuracy for the whole test set, with \textit{Imb. Acc} the top-1 accuracy for the imbalanced classes while also providing the standard deviation for the latter. Best performing combinations are emphasized with bold. Models that were trained using Sefa-based data augmentation tend to outperform both the ones trained on the originally imbalanced datasets and the ones that used baseline, geometric data augmentation methods in most settings or are at least within comparison limits. It is worth noting that when both data augmentation methods are employed (Mixed), the overall performance surges, especially for the imbalanced classes, thus suggesting that the combination of geometric and context-based data augmentation arms the model with more useful variability than when they are employed separately. Experiments that use a larger sample size than the others are denoted by asterisk.\label{tbl:metrics}}
\end{table*}
\subsubsection{Experimental Setting}
\begin{enumerate}
    \item \textit{Classification}: Each of the experiments in this subsection involved fine-tuning a standard ResNet-50 pre-trained on ImageNet on each \{dataset variant, data augmentation strategy\} combination. The duration of the training was fixed to 30 epochs and the optimizer choice was Adam with a learning rate of 0.001. For compatibility purposes, we normalized each input image using Imagenet standards. The batch size we opted for was 512.
    \item \textit{Data Augmentation}: 
    \begin{enumerate}
        \item \textit{Baseline Data Augmentation}: For baseline data augmentation purposes, we employed three distinct rotations, the angles of which were sampled randomly from \{30\degree, 60\degree, 90\degree, 120\degree, 150\degree, 210\degree, 240\degree, 270\degree\}, plus a single horizontal flip, for each imbalanced class training sample of the datasets described in subsection \ref{imbalanced_datasets}, thus leading to an increase in training size by 5.
        \item \textit{SeFa-based Data Augmentation}: For our own, SeFa based, data augmentation strategies we chose to repeatedly: i) generate a novel image via our SWAGAN model ii) generate 4 augmented images by exploration for each such image, iii) pseudo-label all images via our ResNet-50 annotator model, trained on the complete NWPU RESISC-45 dataset, and, iv) sample a subset from $[0,...,4]$ that indicates which of those images are to be included in the final rebalanced dataset, until an adequate number of examples per class was gathered.
    \end{enumerate}
     Every data augmentation strategy (baseline, or SeFa-based) was evaluated using the exact same train/val/test ratios and counts, except for that denoted as \textit{Mixed} where both baseline and SeFa-based augmentations were exploited, thus doubling the data augmented examples that were available.
\end{enumerate} 
\subsubsection{Results and Discussion}
\label{res_and_dis}
Results on the data augmentation for remote scene classification are presented in Table  \ref{tbl:metrics}, while examples of images generated by the proposed framework and used in our augmentation pipeline are presented in Fig. \ref{fig:data_augmentations}.  By observing the results, we can see that the variability introduced by our framework gives rise to higher accuracy across all dataset configurations - except in one case, where the results are still comparable to the improvement brought on by our framework.    The aforementioned observations hold both in terms of overall accuracy, as well as the average accuracy in terms of imbalanced classes. 

Interestingly, while we can expect geometric transformations in the baseline setting to generalize across datasets - as the geometric variability is always useful - we also find that data synthesized via the proposed approach generalize across datasets.  That is, while our pre-trained SWAGAN was trained on the NWPU-RESISC45 dataset, it can still yield consistent improvements when augmenting datasets it has not been trained on - such as the AID and UC-Merced datasets.   This evinces of the generalization properties of the synthesis framework, as the generated samples can be readily used for improving the accuracy of classifiers on unseen datasets.

Furthermore, it should be noted that a noticeable upward trend  can be observed in terms of performance as training size increases (Resisc10 to Resisc35 to Resisc70).

Finally, we highlight that by combining the variability introduced by both baseline augmentations, and the proposed framework ({\it Mixed} setting), provides the most accurate results - which is consistent with our observation regarding the disjoint variability introduced by each of the approaches, and confirms that the two can be used in a {\it synergistic} manner.

\section{Conclusion}
\label{sec:conclusion}

In this work, we presented the first pre-trained style-based GAN for satellite image generation. By leveraging recent advances in wavelet-driven generative models, namely the SWAGAN \cite{gal2021swagan}, we are able to well-capture high-frequency details that naturally arise in satellite imagery, producing a wide gamut of imagery.  Furthermore, we showed that by performing an analysis of the weight-space of the pre-trained GAN results in distinct directions that correspond to high-level semantic concepts that are inherent to the domain of satellite images, including concepts such as {\it urbanization, vegetation growth, wetland density}.  As demonstrated by our experiments, the synthetic imagery is realistic, and can be readily used for downstream tasks such as data augmentation for classification - even on datasets that the generative model has not been trained on.

\section*{Acknowledgements}
\label{sec:acknowledgements}

This work has been partially supported by the Cyprus Research and Innovation foundation under contract number  PRE-SEED/0719/0042.

\bibliographystyle{abbrv}
\bibliography{arxiv}

\end{document}